\documentclass[runningheads]{llncs}
\usepackage[T1]{fontenc}
\usepackage{hyperref}
\usepackage{algorithm}
\usepackage{algpseudocode}
\usepackage{physics}
\usepackage{graphicx}
\usepackage[caption=false]{subfig}
\usepackage{todonotes}
\usepackage{color}

\usepackage{cite}
\usepackage{amsmath, amssymb}
\begin{document}

\title{Physics-Informed Diffusion Models for Unsupervised
Anomaly Detection in Multivariate Time Series}
\titlerunning{Physics-informed Diffusion Model for Anomaly Detection}
%
\author{Juhi Soni\inst{1} \and
Markus Lange-Hegermann\inst{2} \and
Stefan Windmann\inst{1}}

\authorrunning{J. Soni,  M. Lange-Hegermann, S. Windmann}
%
%
\institute{Fraunhofer IOSB-INA, Lemgo, Germany\\
\email{\{juhi.soni,stefan.windmann\}@iosb-ina.fraunhofer.de}\\
\and
TH-OWL University of Applied Sciences and Arts Lemgo, Germany\\
\email{markus.lange-hegermann@th-owl.de}}
\maketitle              
\begin{abstract}
We propose an unsupervised anomaly detection approach based on a physics-informed diffusion model for multivariate time series data. Over the past years, diffusion model has demonstrated its effectiveness in forecasting, imputation, generation, and anomaly detection in the time series domain. In this paper, we present a new approach for learning the physics-dependent temporal distribution of multivariate time series data using a weighted physics-informed loss during diffusion model training. A weighted physics-informed loss is constructed using a static weight schedule. This approach enables a diffusion model to accurately approximate underlying data distribution, which can influence the unsupervised anomaly detection performance. Our experiments on synthetic and real-world datasets show that physics-informed training improves the F1 score in anomaly detection; it generates better data diversity and log-likelihood. Our model outperforms baseline approaches, additionally, it surpasses prior physics-informed work and purely data-driven diffusion models on a synthetic dataset and one real-world dataset while remaining competitive on others.

\keywords{Time Series \and Anomaly Detection \and Physics-informed machine learning \and Diffusion Model.}
\end{abstract}

\section{Introduction}
Anomaly detection techniques play an important
role in various applications. They can be used to identify system
defects, failures, network attacks, and more. In the industry, detecting abnormal patterns in data is crucial for ensuring smooth operations and reducing process downtime \cite{pintilie2023time}. Researchers have introduced many supervised, semi-supervised, and unsupervised anomaly detection techniques over the years, driven by the increasing availability of data. Given the limited requirement for labeled data in semi-supervised methods and the absence of such a requirement in unsupervised methods, research efforts have concentrated on semi-supervised and unsupervised learning approaches\cite{pintilie2023time,hammerbacher2021including}.
The generative models have shown their capability to generate high-quality samples across various domains \cite{kingma2022autoencoding, goodfellow2014generative, ho2020denoising, song2020score, pml2Book}, such as text-to-image generation, audio generation, video generation, etc. In recent years, there has been a growing interest in the application of generative models in time series analysis, particularly for anomaly detection. These models \cite{kingma2022autoencoding, goodfellow2014generative, ho2020denoising, song2020score} have been applied to a range of time series tasks, including forecasting, imputation, synthetic data generation, and anomaly detection. Generative models are used in unsupervised anomaly detection because of their ability to model complex data distributions by learning the underlying structure of normal data and identify anomalies as deviations from the expected patterns. In time series anomaly detection, various variants of Variational Autoencoders (VAEs) \cite{lin2020anomaly, guo2018multidimensional, su2019robust, windmann2024fault}, Generative Adversarial Networks (GANs) \cite{niu2020lstm, li2019mad, jiang2019gan}, diffusion models \cite{pintilie2023time, tian2024prodiffad, sui2024anomaly, hu2024unsupervised}, and auto-regressive models are commonly used. 

In the last few years, the diffusion model has gained popularity due to its ability to generate high-quality samples, optimize log-likelihood, mode coverage, and training stability \cite{ho2020denoising, song2020score, kingma2021variational, vahdat2021score, rombach2022high}. In the time series domain, diffusion model based reconstruction and imputation techniques for anomaly detection are introduced in \cite{pintilie2023time, sui2024anomaly, hu2024unsupervised, xiao2023imputation}. All these anomaly detection approaches rely solely on the data. In domains such as fluid dynamics, biology, and engineering, data inherently adhere to physical laws, and the standard diffusion model training and sampling ignore these underlying physical laws. Therefore, to incorporate physics into the model, in \cite{ shu2023physics}, physics residual is included as conditioning information for high-fidelity flow field reconstruction. Bastek et al. \cite{bastek2024physics} defined a physics-informed loss function using Partial Differential Equation (PDE) and scaled variance of the diffusion process, with a focus on minimizing the physics constraints during model training. All previous works on physics-informed diffusion models \cite{bastek2024physics, shu2023physics, jacobsen2023cocogen, christopher2024projected} are focused on applying PDE constraints during model training or sampling on image datasets.

In the time series domain, the generative model must capture the temporal relationships in addition to the feature distribution \cite{desai2021timevae}. These temporal relationships in some datasets follow the laws of physics. Incorporating prior knowledge improves the model’s fit and enhances its interpretability \cite{besginow2022constraining}. Previous work on diffusion models for time-series anomaly detection has not exploited physical laws during training or inference. Inspired by this recent work on anomaly detection introduced above \cite{pintilie2023time, sui2024anomaly, hu2024unsupervised, xiao2023imputation} and the physics-informed diffusion model \cite{bastek2024physics, shu2023physics}, we propose a temporal physics-informed diffusion model for multi-variant time series anomaly detection. Here, we aim to improve anomaly detection capabilities by exploiting the physical knowledge of the data during model training.

In this paper, we propose a novel approach to add physics information during model training. We define a weight for every diffusion time step and use it to build the physics-informed loss function for physics aware diffusion model training. We design a weighting schedule that reduces the influence of noisy data relative to less noisy data in the construction of the physics-informed loss function. Our approach improves the model's ability to learn the underlying temporal distribution of data, ultimately resulting in improved robustness for anomaly detection. In this work, we have evaluated the proposed anomaly detection method on four multivariate time series datasets in terms of F1 score, log-likelihood, and data diversity. To validate our approach, we assessed the statistical significance of the performance gains achieved through physics-informed training.\\ 

Our contributions are summarized as follows:
\begin{enumerate}  
  \item We develop a temporal physics-informed diffusion model for anomaly detection that adheres to the underlying physical laws governing the data by incorporating weighted physics-informed loss function during diffusion model training.
  \item We define and incorporate a weight schedule into the physics-informed loss function.
  \item  Assessing the physics-informed and uninformed diffusion model by validating them based on the log-likelihood, data diversity, and F1 score.
\end{enumerate}

\section{Background}
\subsection{Evidence Lower Bound (ELBO)}
In unsupervised anomaly detection, we can model the normal behavior of data using a probability distribution by leveraging the Evidence Lower Bound (ELBO) objective, and then define a threshold to distinguish between normal and anomalous data. The ELBO is a lower bound on the marginal likelihood of data $x$ and a good approximation to $\log p(x)$.   Within a variational inference scheme, we introduce latent variables, $z$, sampled from a prior distribution $p(z)$. A simpler distribution, $q(z |x)$, is then used to approximate the true posterior, $p(z | x)$ \cite{blei2017variational, pml2Book}. These posterior approximations enable the model to capture the underlying distribution of normal data, facilitating the more effective identification of deviations or anomalies in the latent space.
\[
\mathbf{ELBO} = \mathbb{E}[\log p(x | z)] - \text{$D_{KL}$}(q(z | x) || p(z | x))
\]
So, by maximizing the ELBO, we ensure the variational distribution $q(z | x)$ closely approximates the true posterior $p(z| x)$ by minimizing their Kullback-Leibler $D_{KL}$ divergence.
\subsection{Denoising Diffusion Probabilistic Models (DDPM)} \label{ddpm}
A diffusion probabilistic model is a latent variable model parameterized by a Markov chain, which is trained using variational inference techniques to generate samples that closely match the data within a finite time frame \cite{ho2020denoising}. The training of this model consists of two phases: a forward diffusion process and a reverse denoising process. In the forward process, the approximate posterior is fixed to a Markov chain that gradually introduces Gaussian noise to the data following a variance schedule \textit{$ \sigma_{1}, ..., \sigma_{T}$} \cite{ho2020denoising}.
The forward process is defined by
\[
  \textit{$q(x_{t}\vert{x_{t-1}})$} = \mathcal{N}(x_{t}; { \sqrt{\alpha_{t}}x_{0}, (1 - \alpha_{t}) I })
\]
where \textit{$\alpha_{t} = 1 - \sigma_{t}$}.
\[
  \textit{$q(x_{t}\vert{x_{0}})$} = \mathcal{N}(x_{t}; { \sqrt{\overline{\alpha_{t}}}x_{0}, (1 - \overline{\alpha_{t}}) I })
\]
where \textit{$\overline{\alpha_{t}} = \prod_{s=1}^{t}\alpha_{s}$}. The reverse diffusion process is a generative process in which we invert the forward diffusion process. It is defined by
\[
  \textit{$p_{\theta}(x_{t-1}\vert{x_{t}})$} = \mathcal{N}(x_{t-1}; { \mu_{\theta} (x_{t}, t),\,\Sigma_{\theta} (x_{t}, t) })\
\]
where \textit{$\Sigma_{\theta} (x_{t}, t) = \sigma_{t}^2 I$} or \textit{$\frac{1 - \overline{\alpha}_{t - 1}}{1 - \overline{\alpha}_{t}} \sigma_{t}$} \cite{ho2020denoising} and the generator $\mu_{\theta} (x_{t}, t)$ approximate the data distribution. The combination of forward process \textit{q} and reverse process \textit{p} constitutes a variational autoencoder \cite{kingma2022autoencoding, kingma2017variational, nichol2021improved}, and the model is trained by maximizing the ELBO. \\
\begin{align*}
\textit{$\mathcal{L_{DM}}$ = $\underbrace{\textit{$ D_{KL}(q(x_{T}\vert{x_{0}}) || p(x_{T})) $}}_{L_{T}}$} \underbrace{\textit{$ + \sum_{t=2}^{T} D_{KL}(q(x_{t-1}\vert{x_{t}, x_{0}}) || p_{\theta}(x_{t-1} \vert x_{t}))$}}_{L_{t-1}} \\ 
-\underbrace{ \log p_{\theta} (x_{0} \vert x_{1})}_{L_{0}}
\end{align*}
The $L_{T}$ and $L_{0}$ computed using standard techniques \cite{kingma2017variational} and since all distributions are Gaussian $L_{t-1}$ can be defined by \cite{ho2020denoising}
\[
  L_{t-1} =  \textit{$ \mathbb{E}_{q} \left [ \frac{1}{2\sigma^2} || \tilde{\mu}(x_{t}, x_{0}) -  \mu_{\theta} (x_{t}, t) ||^2 \right] $} + C
\]
As per  \cite{ho2020denoising}, the model can be parameterized as a noise $\epsilon$ prediction model $\epsilon_{\theta}$ by using the simplified ELBO. 
\[
  \mathcal{L_{DM}} =  \textit{$ \mathbb{E}_{t, x_{0}, \epsilon} \left[ || \epsilon -  \epsilon_{\theta} (\sqrt{\overline{\alpha_{t}}}x_{0} + \sqrt{1 - \overline{\alpha_{t}}} \epsilon, t) ||^2 \right] $}
\]
In addition, the neural network $x_{\theta}$ can also be parameterized to predict clean data $x_{0}$.

\subsection{Physics-informed Neural Network} \label{dr}
Physics-informed Neural Network (PINN) \cite{raissi2017physicsI, raissi2017physicsII} works as a data-driven function approximator, where physical laws are exploited in the training process. Many laws of physics are described by Partial Differential Equations (PDE) or Ordinary Differential Equations (ODE). In Physics-informed Neural Networks, we train a model with the underlying physics knowledge of data by embedding this knowledge into the network's loss function \cite{raissi2017physicsI, misyris2020physics, cuomo2022scientific, meng2025physics}. The overall training objective is a composite loss function, $\mathcal{L}$, which combines a standard network loss, $\mathcal{L_{F}}$, with a physics-informed loss $\mathcal{L_{PI}}$.
\[
  \textit{$\mathcal{L}$} = \textit{$\mathcal{L_{F}}$} + \textit{$\mathcal{L_{PI}}$}
\]
The physics-informed loss, $\mathcal{L_{PI}}$ is defined by
\[
  \textit{$\mathcal{L_{PI}}$} = \textit{$\frac{1}{N} \sum_{i=1}^{N} \mathcal{E}(\mathcal{D}, F(x_i; \theta))^2$ }
\]
where $\mathcal{E}$ denotes the residual of the governing physical equations, $\mathcal{D}$ denotes the physics prior, and $F(x; \theta)$ represents the output of neural network. The physics-informed loss term works as a regularizer that limits model parameters to applied physics law. Encoding physics information into a learning algorithm enhances the informational content of the data, leading to a boost in neural network model performance  \cite{raissi2017physicsI}.

\section{Related Work}
The physics-aware diffusion model was introduced in \cite{jacobsen2023cocogen} to generate samples constrained by partial differential equations (PDE) by minimizing the residue during model sampling. Christopher et al. \cite{christopher2024projected}, augmented diffusion-based synthesis with constraints during sampling to generate high-fidelity content. A physics-guided motion diffusion model (PhysDiff) \cite{yuan2023physdiff} incorporates physical constraints into the diffusion process to generate physically implausible motions. For high-fidelity flow field reconstruction in \cite{shu2023physics}, physics residual is included as conditioning information during model training and inference.
Similar to our approach, \cite{bastek2024physics} introduces a physics-informed residual based on partial differential equations (PDEs) and the scaled variance of the diffusion process. This residual is incorporated into the training of a diffusion model to generate samples that satisfy the underlying PDE. These studies focus on incorporating physics-based constraints into model training or sampling to produce physically consistent samples. Apart from \cite{bastek2024physics} and \cite{yuan2023physdiff}, they overlook the noise components intrinsic to the diffusion process. Our weighting schedule mitigates the impact of noise during physics-informed training, resulting in a closer fit to the data distribution.\\ \\
In time series anomaly detection in \cite{pintilie2023time} the autoencoder model is used with the diffusion model to identify anomalies using a reconstruction signal. A Denoising Diffusion Time Series Anomaly Detection (DDTAD) \cite{sui2024anomaly} also utilizes the reconstruction error for detecting abnormalities. Xiao et al. \cite{xiao2023imputation} have used a generative time series imputation process for anomaly detection. A TimeADDM \cite{hu2024unsupervised} uses multi-level reconstructions of diffusion models for unsupervised anomaly detection for multivariant time series. The ImDiffusion \cite{chen2023imdiffusion} method reduces the uncertainty of data and enhances anomaly detection by employing imputation to integrate information from neighboring time points, effectively capturing temporal and correlated patterns. Analogous to our approach in \cite{zuo2024unsupervised} normal data distribution features are learned during training and abnormal data are detected through sequence reconstruction. The benefit of our work over existing methods lies in the utilization of ELBO for the detection of anomalous patterns, and the integration of a weighted physics law during model training yields a physically consistent model, resulting in a statistically significant improvement in anomaly detection performance.

\section{Temporal Physics-informed Diffusion Model (TPIDM)}
\begin{figure}[ht]
\centering
\includegraphics[width=\textwidth]{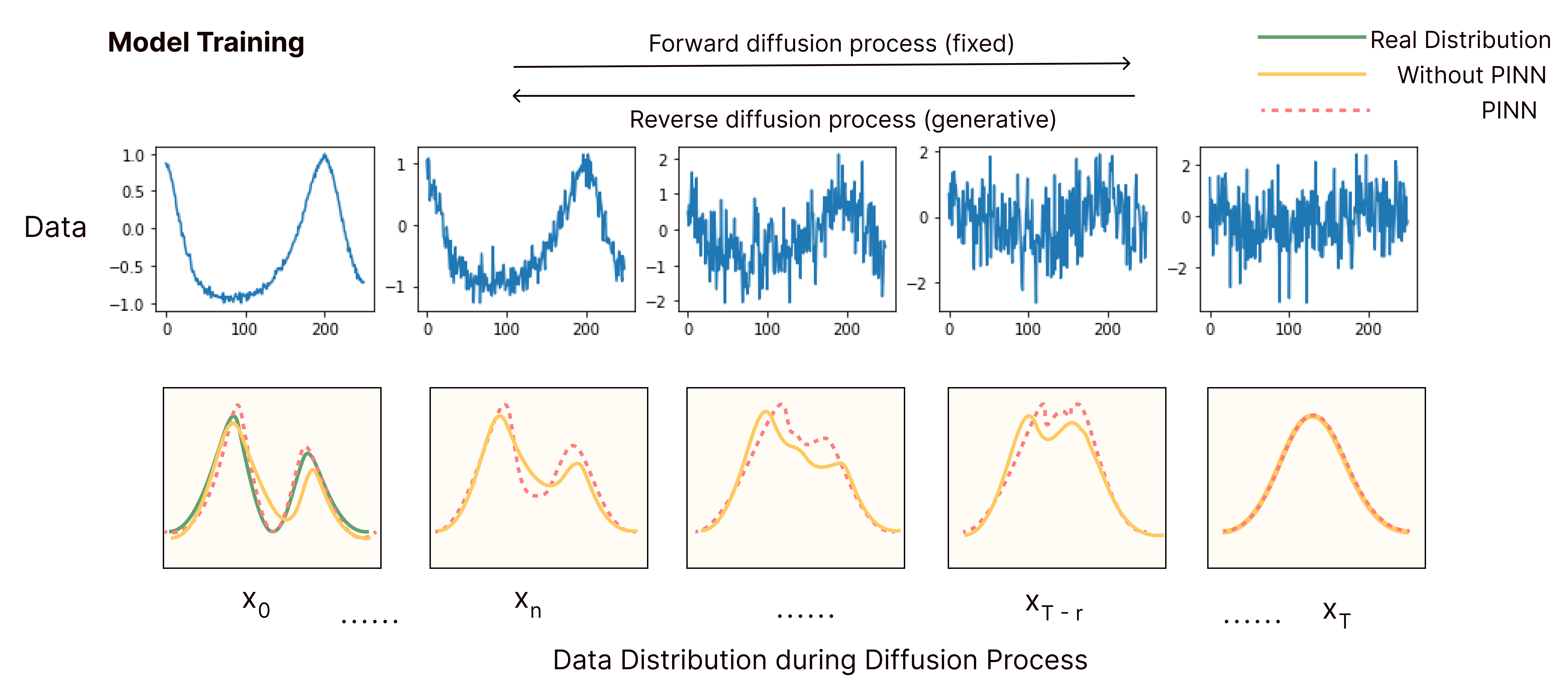}
\caption{Data-driven diffusion model vs physics-informed diffusion model training. A weighted Physics-informed loss is added to the diffusion model training, which leads to an improvement in the model’s ability to approximate the underlying data distribution.}
\label{fig:tpidm-train}
\end{figure}
To learn the physics-dependent temporal distribution of data, the underlying physics of the data is integrated into the loss function of the diffusion model, thereby embedding the physical knowledge of the data into the model's learning process. The physics-informed loss is formulated using the reconstructed signal and the underlying physical laws governing the data. It is evident from Figure \ref{fig:tpidm-train} that the data near the time step \textit{$t = 0$} exhibits significantly lower noise levels, indicating that earlier diffusion steps preserve more of the original signal structure. Furthermore, at earlier diffusion steps, the signal is closer to being a smooth curve; hence, derivatives can be estimated much more reliably. In contrast, when reconstructing the signals $\widehat{x_{0}}$ from $x_{t}$ at diffusion time step $T$ or near $T$, the resulting outputs contains higher noise levels than those reconstructed from earlier time steps, as the model has not yet reached convergence. Due to these noisy reconstructed signals, the physics-informed loss is inadequately defined because of the highly noisy derivative estimations. This adversely affects the model’s learning process, causing divergence from optimal solutions and resulting in anomalous or improbable outputs. So, to address this poorly defined physics-informed loss and enhance the model's performance, we propose a static weight schedule for physics-informed loss functions. The physics-informed loss weight schedule, denoted $\overline{\lambda_{PI_{t}}}$, is constructed to prioritizes toward the initial $n$ diffusion steps by assigning them greater weight in the loss computation, while subsequent steps approaching the final step $T$ are assigned zero weight.
\[
  \mathcal{L_{PI}} = \overline{\lambda_{PI_{t}}} \textit{$\frac{1}{N} \sum_{i=1}^{N} \mathcal{E}(\mathcal{D}, \widehat{x_{i_{0}}})^2$}
\]
Here \textit{$\overline{\lambda_{PI_{t}}} = \prod_{s=1}^{t}\lambda_{PI_{s}}$} and $\lambda_{PI_{s}}$ is defined by \textit{$f(s) \cdot (m - n) + l $}, where $f(s)$ is a nonlinear function, $m$ and $n$ determine the amplitude bounds, $l$ is offset. Here $\overline{\lambda_{PI_{t}}}$ eliminates the noise in the reconstructed signal, $\widehat{x_{i_{0}}}$. This noise which is inherited from signal $x_{i_{t}}$ at an arbitrary diffusion step $t$, and its removal ultimately enhances the reliability of the $\mathcal{L_{PI}}$  computation. This weight schedule is bounded between 0 and 1, and it saturates at 0. Its shape and saturation points depend on the $f(s)$. This weighting schedule ensures that the constructed physics loss is more precise than one computed using all diffusion steps. The Diffusion Model loss function with physics-informed loss is as follows
\[
  \mathcal{L_{TPIDM}} = \mathcal{L}_{DM} + \mathcal{L_{PI}}
\]
where $\mathcal{L}_{DM}$ indicates the diffusion model loss function. 

\subsection{Anomaly Detection}
\begin{figure}[ht]
\centering
\includegraphics[width=\textwidth]{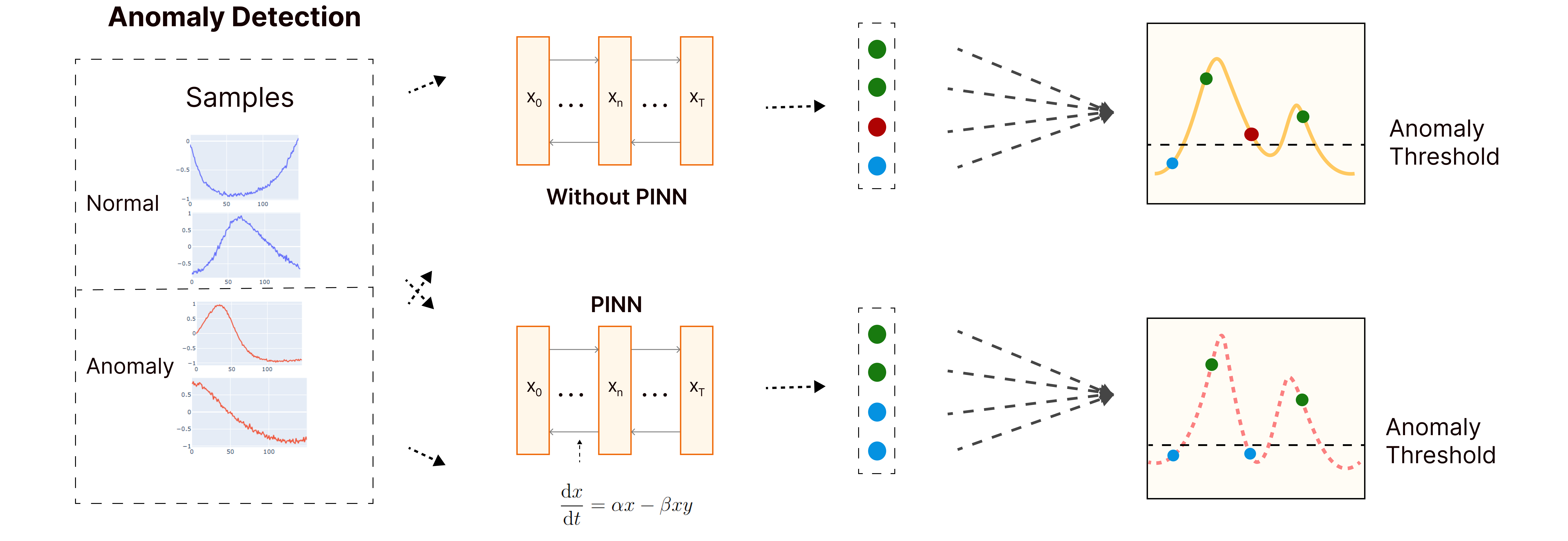}
\caption{ Physics-informed training influences the model’s ability to learn and generalize the underlying data distribution, resulting in the correct identification of normal and anomaly data compared to an uninformed diffusion model.}
\label{fig:tpidm-anomaly}
\end{figure}
The diffusion model learns to approximate the underlying probability distribution of the data. After training the model exclusively on normal data, anomaly detection is performed based on its Evidence Lower Bound (ELBO), which serves as the diffusion model's loss function $\mathcal{L}_{DM}$. As shown in Figure \ref{fig:tpidm-anomaly}, a testing sample is labeled as an anomaly if the computed ELBO of the sample exceeds the threshold. The anomaly threshold is determined using \cite{sui2024anomaly}
\[
  \mathcal{A}_{thr} = \mu_{trim} + k \cdot \operatorname{iqr}
\]
with $\mu_{trim}$ and $\operatorname{iqr}$ denoting the trimmed mean and interquantile range of the model’s loss over the validation data, respectively.
\section{Experiments}
The proposed anomaly detection method has been evaluated using four multivariate time series datasets (three real-world datasets and a synthetic dataset described in Appendix \ref{datasets}). The TPIDM's anomaly detection results are compared against vanilla autoencoder, Variational Autoencoder, K-means algorithm, and two Physics-informed \cite{shu2023physics, bastek2024physics} diffusion models. The diffusion models are further evaluated quantitatively using log-likelihood estimates on the validation dataset and the binary F1 score to assess anomaly classification performance. We use imbalanced data (700 normal and 300 anomalies) to evaluate the performance of a model in anomaly detection. We investigate the statistical significance of performance improvements achieved by the physics‑informed diffusion model over the uninformed diffusion model and baseline methods for anomaly detection. The diversity of the generated data is evaluated by the Principle Component Analysis (PCA) \cite{pml1Book} charts of the original and synthetic data \cite{yoon2019time, desai2021timevae}. Details regarding the model architecture, hyperparameters, model training, and inference time are provided in Appendix \ref{model-architecture}.
\subsection{ELBO and Sample Diversity}
\begin{figure}
\subfloat[Diffusion model without PINN\label{fig:pca-wpinn}]{%
  \includegraphics[width=0.3\textwidth]{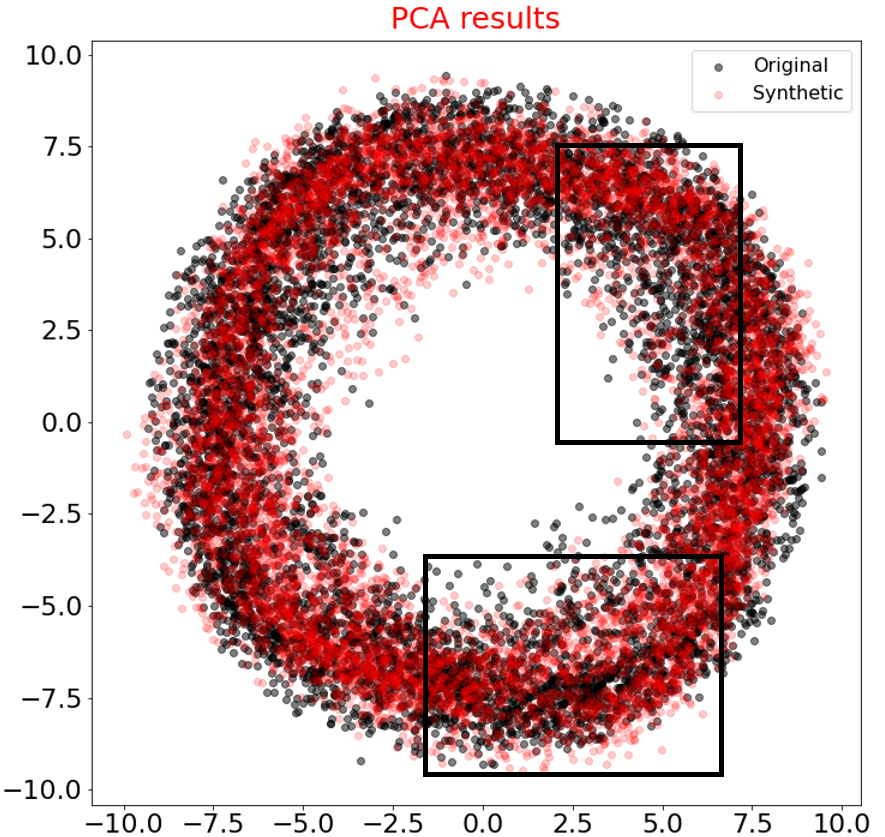}%
}\hfil
\subfloat[PIDM \cite{shu2023physics}\label{fig:pca-pinn-cond}]{%
  \includegraphics[width=0.3\textwidth]{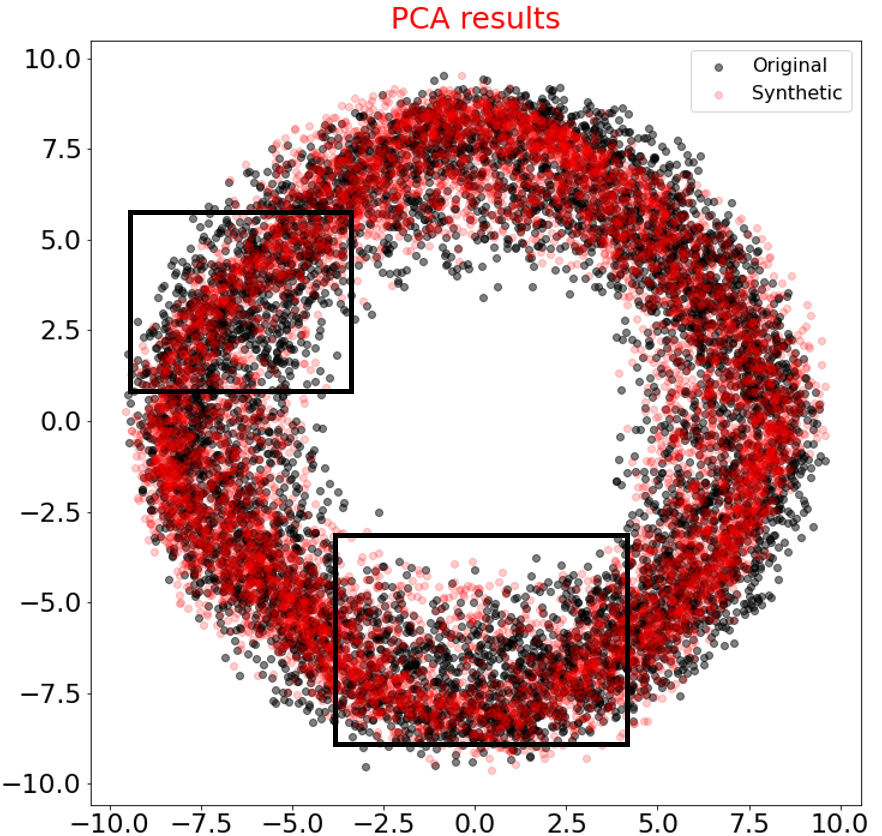}%
}\hfil
\subfloat[TPIDM (ours)\label{fig:pca-ours}]{%
  \includegraphics[width=0.3\textwidth]{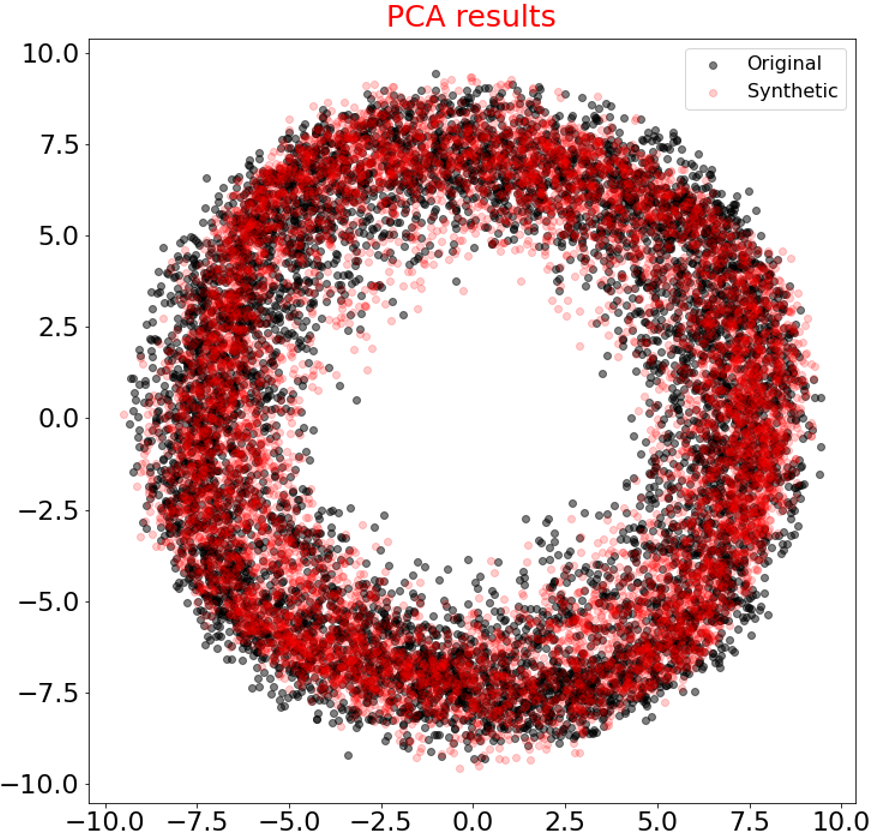}%
}
\caption{PCA plots of generated and original Lenze data. The red portion represents the synthetic data, while the black portion corresponds to the original data. The bounding boxes highlight areas not covered by the generated data in Figures \ref{fig:pca-wpinn} and \ref{fig:pca-pinn-cond}. These plots indicate that our model achieves more overlap between the red and black regions between the synthetic and original data, suggesting improved data diversity.}
\label{fig:data-diversity-results}
\end{figure}

Table \ref{table:dm-elbo-table} shows the impact of our method on the likelihood. We achieved better log-likelihoods across the datasets EMPS, Predator-Prey, and Air Compressor for a noise prediction model ($\epsilon_{\theta}$). The $x_{0}$ prediction model outperformed the other methods on the EMPS, Lenze, and Air Compressor datasets. Prior physics-informed \cite{shu2023physics, bastek2024physics} diffusion models' results are compared based on their diffusion loss ($\epsilon$ or $x_{0}$). Figure \ref{fig:data-diversity-results} illustrates the performance of the models with respect to data diversity. It shows that our method improved the diversity and performed better compared to a data-driven model and the model proposed in \cite{shu2023physics} when trained via noise prediction objective. However, training the model ($x_{\theta}$) using the $x_{0}$ prediction objective did not yield any improvement in sample diversity.
\begin{table}[H]
    \centering        
    \scriptsize
    \caption{ELBO comparison of data-driven diffusion model (DM without PINN), prior physics-informed diffusion model introduced in \cite{shu2023physics, bastek2024physics} (PIDM), and our method.}
    \begin{tabular}{c |c | c | c | c }
        \hline
           ~ & \multicolumn{4}{c}{\textbf{ELBO}} \\ \hline
          &  \textbf{EMPS} & \textbf{Predator-Prey} & \textbf{Lenze} & \textbf{Air Compressor}\\ \hline         
          \textbf{DM ($\epsilon_{\theta}$)} & -703.56 $\pm$ 0.45 &  -612.46 $\pm$ 0.12 & -4779.47 $\pm$ 3.13 & -4224.99 $\pm$ 2.47\\          
          \textbf{PIDM} \cite{shu2023physics} & -628.46 $\pm$ 2.02 & -591.85 $\pm$ 0.86 & \textbf{-4816.08} $\pm$ 1.26 & -3993.55 $\pm$ 11.57\\
          \textbf{TPIDM (Ours, $\epsilon_{\theta}$)} & \textbf{-707.99 $\pm$ 0.81} & \textbf{-613.15 $\pm$ 0.00} & -4767.04  $\pm$ 3.39 & \textbf{-4301.11 $\pm$ 4.34}\\
          \hline
          \textbf{DM ($x_{\theta}$)} & -819.21 $\pm$ 0.69 &  \textbf{-490.87 $\pm$ 0.00} & -6291.91 $\pm$ 0.00 & -4061.78 $\pm$ 0.00\\      
          \textbf{PIDM} \cite{bastek2024physics} & 28989.58 $\pm$ 3075.11 & 708.47  $\pm$ 0.04 & -5289.29 $\pm$ 0.01 & -761.68 $\pm$ 121.41\\
          \textbf{TPIDM (Ours, $x_{\theta}$)} & \textbf{-821.78 $\pm$ 0.02} & -469.81 $\pm$ 0.00 & \textbf{-6303.48 $\pm$ 0.00} & \textbf{-4111.57 $\pm$ 2.02}\\
          \hline
    \end{tabular}
    \label{table:dm-elbo-table}
\end{table}

\subsection{Anomaly Detection Results}
\begin{table}[H]
    \centering        
    \scriptsize
    \caption{F1 score comparison of AE, VAE, K-means, data-driven diffusion model (without PINN), prior physics-informed diffusion model \cite{shu2023physics, bastek2024physics}, and our method}
    \begin{tabular}{c |c  c  c  c }
        \hline
           ~ & \multicolumn{4}{c}{\textbf{F1 Score}} \\ \hline
          &  \textbf{EMPS} & \textbf{Predator-Prey} & \textbf{Lenze} & \textbf{Air Compressor} \\ \hline
          Other models \\
          \textbf{VAE} & 0.9624 $\pm$ 0.0047 & 0.9133 $\pm$ 0.0011 & 0.8298 $\pm$  0.0031 & 0.9664 $\pm$ 0.0389\\
          \textbf{AE} & 0.9168 $\pm$ 0.0030 & 0.8226 $\pm$ 0.0000 & 0.9820 $\pm$ 0.0000 & 0.9848 $\pm$ 0.0000\\
          \textbf{K-MEANS} & 0.8526 $\pm$ 0.0000 & 0.7868 $\pm$ 0.0000 & 0.8102 $\pm$ 0.0000 & 0.8235 $\pm$ 0.0000\\
          \hline      
          DM with $\epsilon_{\theta}$ \\
          \textbf{DM ($\epsilon_{\theta}$)} & 0.9627 $\pm$ 0.0031 &  0.9712 $\pm$ 0.0017 & \textbf{0.9923 $\pm$ 0.0003} & 1.0 $\pm$ 0.0000\\
          \textbf{PIDM} \cite{shu2023physics} & 0.9606 $\pm$ 0.0067 & 0.9586 $\pm$ 0.0026 & 0.9405 $\pm$ 0.0167 & 1.0 $\pm$ 0.0000\\
          \textbf{TPIDM (Ours, $\epsilon_{\theta}$)} & \textbf{0.9699 $\pm$ 0.0028} & \textbf{0.9830 $\pm$ 0.0000} & 0.9622 $\pm$ 0.0116 & 1.0 $\pm$ 0.0000\\
          \hline
          DM with $x_{\theta}$ \\
          \textbf{DM ($x_{\theta}$)} & \textbf{1.0 $\pm$ 0.0000} &  0.9957 $\pm$ 0.0000 & \textbf{0.9993 $\pm$ 0.0000} & 1.0 $\pm$ 0.0000\\      
          \textbf{PIDM} \cite{bastek2024physics} & 0.85 $\pm$ 0.0014 & \textbf{1.0 $\pm$ 0.0000} & 0.9624 $\pm$  0.0000  & 1.0 $\pm$ 0.0000\\
          \textbf{TPIDM (Ours, $x_{\theta}$)} & \textbf{1.0 $\pm$ 0.0000} & \textbf{1.0 $\pm$ 0.0000} & \textbf{0.9993 $\pm$ 0.0000} & 1.0 $\pm$ 0.0000\\
          \hline
    \end{tabular}
    \label{table:dm-f1-table}
\end{table}
Table \ref{table:dm-f1-table} shows the effectiveness of our method in detecting anomalies. It demonstrates that our method improves the F1 score and outperforms other methods for EMPS and Predator-Prey datasets when a $\epsilon$ prediction model ($\epsilon_{\theta}$) is used for training. The Table \ref{table:dm-f1-table} also reveals that the proposed physics-informed training approach results in higher F1 scores than the models proposed in previous work \cite{bastek2024physics, shu2023physics}. Our method also improves the F1 score on the Predatory-Prey dataset when training an $x_{0}$ prediction model ($x_{\theta}$). The $x_{\theta}$ model demonstrates competitive performance against DM and PIDM \cite{bastek2024physics} models as a result of excluding highly noisy components from the PINN loss, while the DM loss remains fixed throughout the training process, thereby limiting its impact. These results indicate that physics-informed training leads to more accurate models for anomaly detection. Furthermore, these results show that the diffusion model performs better in anomaly detection compared to the other models. Our results demonstrate a statistically significant effect of the physics-informed training in anomaly detection, with a $p$-value less than 0.01 with the Wilcoxon test when compared with prior work and a purely data-driven approach with ten repetitions.
\subsubsection{Ablations}
\begin{table}[H]
    \centering        
    \scriptsize
    \caption{F1 score for different physics-informed schedules: The rows correspond to the F1 score of different PINN weight schedules. These numbers show the superior and competitive performance of physics-informed training in anomaly detection.}
\begin{tabular}{c |c | c | c | c }
        \hline
           ~ & \multicolumn{4}{c}{\textbf{F1 Score}} \\ \hline
          &  \textbf{EMPS} & \textbf{Predator-Prey} & \textbf{Lenze} & \textbf{Air Compressor}\\ \hline   
          \textbf{DM ($\epsilon_{\theta}$)} & 0.9627 $\pm$ 0.0031 &  0.9712 $\pm$ 0.0017 & \textbf{0.9923 $\pm$ 0.0003} & 1.0 $\pm$ 0.0000\\ 
          \textbf{Log-Sigmoid ($\epsilon_{\theta}$)} & \textbf{0.9699 $\pm$ 0.0028} & 0.9738 $\pm$ 0.0017 & 0.9622 $\pm$ 0.0116 & 1.0 $\pm$ 0.0000\\
          \textbf{Hard-Sigmoid ($\epsilon_{\theta}$)} & 0.9697 $\pm$ 0.0025 &  0.9552 $\pm$ 0.0027 & 0.9578 $\pm$ 0.0152 & 1.0 $\pm$ 0.0000\\
          \textbf{Sigmoid ($\epsilon_{\theta}$)} & 0.9447 $\pm$ 0.0082 & \textbf{0.9830 $\pm$ 0.0000} & 0.9104 $\pm$ 0.0115 & 1.0 $\pm$ 0.0000\\
          \textbf{ReLU ($\epsilon_{\theta}$)} & 0.9015 $\pm$ 0.0000 & 0.9749 $\pm$ 0.0017 & 0.9678 $\pm$ 0.0089 & 1.0 $\pm$ 0.0000\\
          \hline
          \textbf{DM ($x_{\theta}$)} & 1.0 $\pm$ 0.0000 &  0.9957 $\pm$ 0.0000 & \textbf{0.9993 $\pm$ 0.0000} & 1.0 $\pm$ 0.0000\\ 
          \textbf{Log-Sigmoid ($x_{\theta}$)} & 1.0 $\pm$ 0.0000 & 0.9887 $\pm$ 0.0000 & 0.9985 $\pm$ 4.71$e^{-7}$ & 1.0 $\pm$ 0.0000\\
          \textbf{Hard-Sigmoid ($x_{\theta}$)} & 1.0 $\pm$ 0.0000 & 0.9971 $\pm$ 0.0000 & 0.9979 $\pm$ 0.0000 & 1.0 $\pm$ 0.0000 \\
          \textbf{Sigmoid ($x_{\theta}$)} & 1.0 $\pm$ 0.0000 & \textbf{1.0 $\pm$ 0.0000} & 0.9986 $\pm$ 0.0000 & 1.0 $\pm$ 0.0000\\
          \textbf{ReLU ($x_{\theta}$)} & 1.0 $\pm$ 0.0000 & 0.9971 $\pm$ 0.0000 & \textbf{0.9993 $\pm$ 0.0000} & 1.0 $\pm$ 0.0000\\
          \hline
    \end{tabular}
    \label{table:dm-ablation-table}
\end{table}
We investigated the effect of various physics-informed weight schedules on model performance. The anomaly detection results for different schedules are shown in Table \ref{table:dm-ablation-table}. The results shown in the table demonstrate that the physics-informed model trained with a logarithmic sigmoid schedule exhibits superior overall performance in comparison to other PINN weight schedules for a $\epsilon_{\theta}$ diffusion model. For the Lenze dataset, we did not find the physics-informed training beneficial in the context of the anomaly detection with ($\epsilon_{\theta}$) model. We also explored a few other PINN weight schedules to improve the results for Lenze datasets but did not find them useful. We tried sine and tanhshrink functions in building the PINN weight schedule and trained the diffusion model with these schedules, which improved the F1 score to 0.98. However, these functions did not outperform the other PINN schedules and, in fact, led to a deterioration in performance, reducing the F1 score to 0.94 on the EMPS dataset. On the other hand, the ReLU schedule overall performs competitively compared to other schedules for a $x_{\theta}$ model, and a sigmoid schedule improved the model's performance in identifying anomalies for the Predator-Prey dataset.
\section{Conclusion}
We introduced a novel approach to learn the physics-dependent temporal relationship of data using a weighted physics-informed loss in diffusion model training. In particular, we developed a  PINN weighting schedule and embedded it within the physics-informed loss formulation. We demonstrated the effectiveness of the proposed method in terms of unsupervised anomaly detection, log-likelihood, and data diversity. Our experimental results indicate that physics-informed training results in an improved F1 score for anomaly detection. Our findings further confirm that with physics-informed training, we achieve better estimation of the data distribution, leading to an improved log-likelihood. Optimizing with the noise prediction objective enhances the diversity of generated samples compared to the purely data-driven model and prior physics-informed work. It shows that our approach performs better than the previous methods by using a novel weighted physics-informed loss scheduled across diffusion steps. Our approach broadly applies to time series governed by physical principles and can be extended to domains such as fluid dynamics and medical imaging, where physical constraints are relevant. One limitation of the model is its longer inference time compared to methods such as AE and VAE, due to its dependence on the number of diffusion time steps.

\bibliographystyle{splncs04}
\bibliography{ref.bib}

\begin{thebibliography}{10}
\providecommand{\url}[1]{\texttt{#1}}
\providecommand{\urlprefix}{URL }
\providecommand{\doi}[1]{https://doi.org/#1}

\bibitem{emps}
A.~Janot, M.G., Brunot, M.: Data set and reference models of emps. In: Nonlinear System Identification Benchmarks (2019)

\bibitem{bastek2024physics}
Bastek, J.H., Sun, W., Kochmann, D.M.: Physics-informed diffusion models. arXiv preprint arXiv:2403.14404  (2024)

\bibitem{besginow2022constraining}
Besginow, A., Lange-Hegermann, M.: Constraining gaussian processes to systems of linear ordinary differential equations. Advances in Neural Information Processing Systems  \textbf{35},  29386--29399 (2022)

\bibitem{blei2017variational}
Blei, D.M., Kucukelbir, A., McAuliffe, J.D.: Variational inference: A review for statisticians. Journal of the American statistical Association  \textbf{112}(518),  859--877 (2017)

\bibitem{chen2023imdiffusion}
Chen, Y., Zhang, C., Ma, M., Liu, Y., Ding, R., Li, B., He, S., Rajmohan, S., Lin, Q., Zhang, D.: Imdiffusion: Imputed diffusion models for multivariate time series anomaly detection. arXiv preprint arXiv:2307.00754  (2023)

\bibitem{christopher2024projected}
Christopher, J.K., Baek, S., Fioretto, F.: Projected generative diffusion models for constraint satisfaction. arXiv preprint arXiv:2402.03559  (2024)

\bibitem{cuomo2022scientific}
Cuomo, S., di~Cola, V.S., Giampaolo, F., Rozza, G., Raissi, M., Piccialli, F.: Scientific machine learning through physics-informed neural networks: Where we are and what's next (2022)

\bibitem{desai2021timevae}
Desai, A., Freeman, C., Wang, Z., Beaver, I.: Timevae: A variational auto-encoder for multivariate time series generation. arXiv preprint arXiv:2111.08095  (2021)

\bibitem{goodfellow2014generative}
Goodfellow, I.J., Pouget-Abadie, J., Mirza, M., Xu, B., Warde-Farley, D., Ozair, S., Courville, A., Bengio, Y.: Generative adversarial networks (2014)

\bibitem{guo2018multidimensional}
Guo, Y., Liao, W., Wang, Q., Yu, L., Ji, T., Li, P.: Multidimensional time series anomaly detection: A gru-based gaussian mixture variational autoencoder approach. In: Asian Conference on Machine Learning. pp. 97--112. PMLR (2018)

\bibitem{hammerbacher2021including}
Hammerbacher, T., Lange-Hegermann, M., Platz, G.: Including sparse production knowledge into variational autoencoders to increase anomaly detection reliability. In: 2021 IEEE 17th International Conference on Automation Science and Engineering (CASE). pp. 1262--1267. IEEE (2021)

\bibitem{ho2020denoising}
Ho, J., Jain, A., Abbeel, P.: Denoising diffusion probabilistic models. Advances in neural information processing systems  \textbf{33},  6840--6851 (2020)

\bibitem{hoppensteadt2006predator}
Hoppensteadt, F.: Predator-prey model. Scholarpedia  \textbf{1}(10), ~1563 (2006)

\bibitem{hu2024unsupervised}
Hu, R., Yuan, X., Qiao, Y., Zhang, B., Zhao, P.: Unsupervised anomaly detection for multivariate time series using diffusion model. In: ICASSP 2024-2024 IEEE International Conference on Acoustics, Speech and Signal Processing (ICASSP). pp. 9606--9610. IEEE (2024)

\bibitem{jacobsen2023cocogen}
Jacobsen, C., Zhuang, Y., Duraisamy, K.: Cocogen: Physically-consistent and conditioned score-based generative models for forward and inverse problems. arXiv preprint arXiv:2312.10527  (2023)

\bibitem{jiang2019gan}
Jiang, W., Hong, Y., Zhou, B., He, X., Cheng, C.: A gan-based anomaly detection approach for imbalanced industrial time series. IEEE Access  \textbf{7},  143608--143619 (2019)

\bibitem{kingma2021variational}
Kingma, D., Salimans, T., Poole, B., Ho, J.: Variational diffusion models. Advances in neural information processing systems  \textbf{34},  21696--21707 (2021)

\bibitem{kingma2022autoencoding}
Kingma, D.P., Welling, M.: Auto-encoding variational bayes (2022)

\bibitem{kingma2017variational}
Kingma, D.P., et~al.: Variational inference \& deep learning: A new synthesis (2017)

\bibitem{li2019mad}
Li, D., Chen, D., Jin, B., Shi, L., Goh, J., Ng, S.K.: Mad-gan: Multivariate anomaly detection for time series data with generative adversarial networks. In: International conference on artificial neural networks. pp. 703--716. Springer (2019)

\bibitem{lin2020anomaly}
Lin, S., Clark, R., Birke, R., Sch{\"o}nborn, S., Trigoni, N., Roberts, S.: Anomaly detection for time series using vae-lstm hybrid model. In: ICASSP 2020-2020 IEEE International Conference on Acoustics, Speech and Signal Processing (ICASSP). pp. 4322--4326. Ieee (2020)

\bibitem{meng2025physics}
Meng, C., Griesemer, S., Cao, D., Seo, S., Liu, Y.: When physics meets machine learning: A survey of physics-informed machine learning. Machine Learning for Computational Science and Engineering  \textbf{1}(1),  1--23 (2025)

\bibitem{misyris2020physics}
Misyris, G.S., Venzke, A., Chatzivasileiadis, S.: Physics-informed neural networks for power systems. In: 2020 IEEE Power \& Energy Society General Meeting (PESGM). pp.~1--5. IEEE (2020)

\bibitem{MUELLER2024107696}
Mueller, P.N.: Attention-enhanced conditional-diffusion-based data synthesis for data augmentation in machine fault diagnosis. Engineering Applications of Artificial Intelligence  \textbf{131},  107696 (2024). \doi{https://doi.org/10.1016/j.engappai.2023.107696}

\bibitem{pml1Book}
Murphy, K.P.: Probabilistic Machine Learning: An introduction. MIT Press (2022), \url{probml.ai}

\bibitem{pml2Book}
Murphy, K.P.: Probabilistic Machine Learning: Advanced Topics. MIT Press (2023), \url{http://probml.github.io/book2}

\bibitem{nichol2021improved}
Nichol, A.Q., Dhariwal, P.: Improved denoising diffusion probabilistic models. In: International Conference on Machine Learning. pp. 8162--8171. PMLR (2021)

\bibitem{niu2020lstm}
Niu, Z., Yu, K., Wu, X.: Lstm-based vae-gan for time-series anomaly detection. Sensors  \textbf{20}(13), ~3738 (2020)

\bibitem{pintilie2023time}
Pintilie, I., Manolache, A., Brad, F.: Time series anomaly detection using diffusion-based models. In: 2023 IEEE International Conference on Data Mining Workshops (ICDMW). pp. 570--578. IEEE (2023)

\bibitem{raissi2017physicsI}
Raissi, M., Perdikaris, P., Karniadakis, G.E.: Physics informed deep learning (part i): Data-driven solutions of nonlinear partial differential equations. arXiv preprint arXiv:1711.10561  (2017)

\bibitem{raissi2017physicsII}
Raissi, M., Perdikaris, P., Karniadakis, G.E.: Physics informed deep learning (part ii): Data-driven discovery of nonlinear partial differential equations. arXiv preprint arXiv:1711.10566  (2017)

\bibitem{rombach2022high}
Rombach, R., Blattmann, A., Lorenz, D., Esser, P., Ommer, B.: High-resolution image synthesis with latent diffusion models. In: Proceedings of the IEEE/CVF conference on computer vision and pattern recognition. pp. 10684--10695 (2022)

\bibitem{shu2023physics}
Shu, D., Li, Z., Farimani, A.B.: A physics-informed diffusion model for high-fidelity flow field reconstruction. Journal of Computational Physics  \textbf{478},  111972 (2023)

\bibitem{sidebotham2022compressors}
Sidebotham, G.: Compressors and the ideal gas. In: An Inductive Approach to Engineering Thermodynamics, pp. 301--353. Springer (2022)

\bibitem{song2020score}
Song, Y., Sohl-Dickstein, J., Kingma, D.P., Kumar, A., Ermon, S., Poole, B.: Score-based generative modeling through stochastic differential equations. arXiv preprint arXiv:2011.13456  (2020)

\bibitem{su2019robust}
Su, Y., Zhao, Y., Niu, C., Liu, R., Sun, W., Pei, D.: Robust anomaly detection for multivariate time series through stochastic recurrent neural network. In: Proceedings of the 25th ACM SIGKDD international conference on knowledge discovery \& data mining. pp. 2828--2837 (2019)

\bibitem{sui2024anomaly}
Sui, J., Yu, J., Song, Y., Zhang, J.: Anomaly detection for telemetry time series using a denoising diffusion probabilistic model. IEEE Sensors Journal  (2024)

\bibitem{tian2024prodiffad}
Tian, F., Shi, X., Zhou, L., Chen, L., Ma, C., Zhu, W.: Prodiffad: Progressively distilled diffusion models for multivariate time series anomaly detection in jointcloud environment. In: 2024 International Joint Conference on Neural Networks (IJCNN). pp.~1--8. IEEE (2024)

\bibitem{vahdat2021score}
Vahdat, A., Kreis, K., Kautz, J.: Score-based generative modeling in latent space. Advances in neural information processing systems  \textbf{34},  11287--11302 (2021)

\bibitem{windmann2024fault}
Windmann, S., Westerhold, T.: Fault detection in automated production systems based on a long short-term memory autoencoder. at-Automatisierungstechnik  \textbf{72}(1),  47--58 (2024)

\bibitem{Wissbrock2025}
Wißbrock, P., Müller, P.N.: Lenze motor bearing fault dataset (lenze-mb) (2025). \doi{10.5281/zenodo.14762423}, \url{https://doi.org/10.5281/zenodo.14762423}

\bibitem{xiao2023imputation}
Xiao, C., Gou, Z., Tai, W., Zhang, K., Zhou, F.: Imputation-based time-series anomaly detection with conditional weight-incremental diffusion models. In: Proceedings of the 29th ACM SIGKDD Conference on Knowledge Discovery and Data Mining. pp. 2742--2751 (2023)

\bibitem{yoon2019time}
Yoon, J., Jarrett, D., Van~der Schaar, M.: Time-series generative adversarial networks. Advances in neural information processing systems  \textbf{32} (2019)

\bibitem{yuan2023physdiff}
Yuan, Y., Song, J., Iqbal, U., Vahdat, A., Kautz, J.: Physdiff: Physics-guided human motion diffusion model. In: Proceedings of the IEEE/CVF International Conference on Computer Vision. pp. 16010--16021 (2023)

\bibitem{zuo2024unsupervised}
Zuo, H., Zhu, A., Zhu, Y., Liao, Y., Li, S., Chen, Y.: Unsupervised diffusion based anomaly detection for time series. Applied Intelligence  \textbf{54}(19),  8968--8981 (2024)

\end{thebibliography}

\appendix
\section{ Data Sets}\label{datasets}

\subsection{Lotka–Voterra Predator-Prey Dataset}\label{prey-pred}
We generate 100,000 samples of a Lotka–Volterra predator-prey \cite{hoppensteadt2006predator} model with an initial value (x, y) $=$ (10, 2), where x and y denote the population size of preys and predators, respectively. This non-linear biological model describes the interaction between the two species \textit{Predator} and \textit{Prey}. This data set has two features; one describes the growth rate of Prey, and the other describes the growth rate of Predator. 
\begin{multicols}{2}
\ \textbf{Prey Growth Rate} \[ \textit{$ \dv{x}{t} = \alpha x - \beta xy $} \] \ \textbf{Predator Growth Rate} \[ \textit{$ \dv{y}{t} = \delta x y - \gamma y $} \]
\end{multicols}
Here $\alpha$ describes the maximum growth of prey and $\beta$ describes the effect of the presence of predator on the growth rate of prey. A $\delta$ describes the death rate of predator and $\gamma$ describes the effect of the presence of prey on the growth of predator. The values of $\alpha$, $\beta$, $\delta$, and $\gamma$ are 1.1, 0.4, 0.4, and 0.1. These parameters are modified to generate anomalous data.
\subsection{Lenze Dataset}\label{data-lenze}
The Lenze dataset \cite{Wissbrock2025, MUELLER2024107696} is an industrial dataset that is gathered from a three-phase motor. The dataset comprises 524,289 data points and eight features, with the first six channels being related to the electric currents and voltages of the individual phases. We define physics-informed loss using the first six channels by leveraging Ohm's law, which establishes the relationship between voltage, current, and resistance in an electric circuit. The signals originating from the damaged bearing are classified as anomalies.
\[ \dv{V}{t}   = R  \dv{I}{t} \quad\]

\subsection{Electro-Mechanical Positioning
System (EMPS)}\label{emps-data}
This dataset contains three features with 24841 datapoints, respectively. The EMPS is a standard configuration of a drive system for prismatic joint of robots or machine tools \cite{emps}. In the Inverse Dynamic Model (IDM) of a robot, the joint torque or force $\tau$ is expressed as a function of the joint position $q$, velocity, and acceleration \cite{emps}. Here, anomalous data are synthesized by scaling the amplitude of the signals. \[ \tau   = M  \dv[2]{q}{t} + F_{v} \dv{q}{t} + F_c \operatorname{sign}(\dv{q}{t}) + \operatorname{offset}\quad\]
\subsection{Air Compressor Dataset}\label{data-air-comp}
The air compressor dataset is an industrial dataset that comprises 261,640 data points and eight features: Volume, Pressure, Temperature, Mass, Mass Flow Rate, Volumetric Flow Rate, Flow Velocity, and Energy. We compute physics-informed loss based on the ideal law \cite{sidebotham2022compressors}, which establishes the relationship between volume ($v = V/m$), pressure, and temperature.
\[
\textit{$ {P\dv{v}{t}} + {v\dv{P}{t}} = R\dv{T}{t} $}
\]
and the relationship between mass flow rate and volumetric flow rate $ \dot{m} = \rho Q $. Here, leakage occurring in the air compressor system is considered an anomaly.
\section{ Model architecture and Hyperparameters}\label{model-architecture}
We constructed the diffusion model based on an encoder-decoder architecture. We used Long Short-Term Memory (LSTM) networks and SiLU activation function in the encoder-decoder network. The sliding window stride is 1; the number of LSTM layers and hidden neurons, sliding window size, learning rate, and batch size depend on the data set. All models were trained only on normal data. The data was scaled to the interval [-1, 1], and a 90:10 train-test split was used to evaluate for overfitting. Model training was performed using the Adam optimizer with a learning rate of 0.0001 and L2 regularization of \textit{$1e^{-6}$}. For model training, diffusion time steps of T=100 are used. \\ \\
The variances of the forward process are set according to a linear schedule \cite{ho2020denoising} from $\sigma_{1} = 10^{-4}$ to $\sigma_{T} = 0.05$. The variational autoencoder and autoencoder model constructed with the same architecture as the diffusion model. The hyperparameters of the trained models have been optimized in informal experiments, resulting in the values given in Table \ref{table:hype-param}. The parameters of the PINN weight schedule are specified in Table \ref{table:pinn-weight-schedule}\\ \\ \\
\begin{minipage}{0.55\textwidth}
The physics-informed diffusion model and other physics-informed methods are trained with the same architecture and hyperparameter as the uninformed model for every dataset. The time derivative for the physics-informed neural network is computed using finite-difference approximations.
\end{minipage}
\begin{minipage}{0.45\textwidth}
\vspace{-10mm}
\begin{table}[H]
    \scriptsize
    \centering
    \caption{PINN weight schedule}
    \begin{tabular}{c c c c }
        \hline    
        \textbf{\footnotesize{\textbf{f(t)}}} & \footnotesize{\textbf{m}} & \footnotesize{\textbf{n}} & \footnotesize{\textbf{l}} \\ \hline
        Log-Sigmoid & 0.01 & 0.1  & 0.1\\
        Hard-Sigmoid & 0.01  & 1  & 1\\
        Sigmoid & 0.01  & 1  & 1\\        
        ReLU & 0.001 & 0.01 & 0.9\\
        \hline
         \\
    \end{tabular}
    \label{table:pinn-weight-schedule}
\end{table}
\end{minipage}
 \begin{table}[H]
    \scriptsize
    \centering
    \caption{Hyperparameters of all datasets}
    \begin{tabular}{c c c c c }
        \hline    
        \textbf{} & \footnotesize{\textbf{EMPS}} & \footnotesize{\textbf{Predator-Prey}} & \footnotesize{\textbf{Lenze}} & \footnotesize{\textbf{Air Compressor}} \\ \hline
        Batch size & 128  & 128  & 256 & 128\\
        Epochs & 350  & 500  & 500 & 200\\
        Sliding Window & 150  & 100  & 250 & 200\\        
        $c$ of \cite{bastek2024physics} & $10e^{-3}$ & $10e^{-3}$ & $10e^{-3}$ & $10e^{-3}$\\
        Encoder Layers & 3 & 3 & 3 & 3\\        
        Decoder Layers & 3 & 3 & 3 & 3\\
        Encoder Hidden Neurons & (3, 12) (12, 64)  & (2, 8) (8, 16) & (8, 32) (32, 64) & (8, 32) (32, 64) \\
        & (64, 128) & (16, 32) & (64, 128) & (64, 128)\\ 
        Decoder Hidden Neurons & (128, 64) (64, 12) & (32, 16) (16, 8)  & (128, 64) (64, 32) & (128, 64) (64, 32) \\
        &  (12, 3) & (8, 2) & (32, 8) & (32, 8)\\
        Neurons in the latent space & 128 & 32 & 128 & 128\\
        K Means Clusters & 8 & 4 & 16 & 12\\
        \hline
    \end{tabular}
    \label{table:hype-param}
\end{table}
\subsection{ Model Training and Inference Time}\label{model-train-inf}
The training and inference times of the models are shown in
Table \ref{table:train-inference-time}. Here, inference time refers to the duration required by the model to detect anomalies in 1000 (700 normal and 300 anomalies) data points. The training time of the model varies depending on factors such as the dataset size, window length, and number of training epochs. The inclusion of the informed loss function results in a moderate increase in training time, due to the added computation of the PINN loss. The autoencoder achieved the lowest inference time among all evaluated models. Due to the diffusion process, the inference speed of the diffusion model is slower than AE and VAE. 
\begin{table}[ht]
    \centering
    \caption{Training and Inference Time of all datasets}
    \begin{tabular}{c c c c c }
        \hline
        \textbf{} & \footnotesize{\textbf{EMPS}} & \footnotesize{\textbf{Predator-Prey}} & \footnotesize{\textbf{Lenze}} & \footnotesize{\textbf{Air Compressor}} \\ \hline  
        ~ & \multicolumn{3}{c}{\textbf{Training Time}} \\ \hline  \textbf{DM} & 50.20 mins & 5.25 hrs & 1.02 day & 4.42 hrs\\
        \textbf{TPIDM} & 52.87 mins & 5.36 hrs & 1.06 day & 4.59 hrs\\        
        \textbf{PIDM} \cite{bastek2024physics} & 40.64 mins & 5.55 hrs & 1.06 day & 4.33 hrs \\
        \textbf{PIDM} \cite{shu2023physics} & 67.2 mins & 7.02 hrs & 1.26 day & 5.06 hrs\\
        \textbf{VAE} & 49.47 mins & 4.05 hrs & 1.30 day & 5.03 hrs\\
        \textbf{AE} & 47.81 mins & 1.91 hrs & 0.47 day & 3.65 hrs\\
        \textbf{K-MEANS } & 0.40  mins & 0.03 hrs & 0.19 day & 0.041 hrs\\
        \hline
        ~ & \multicolumn{3}{c}{\textbf{Inference Time in seconds}} \\ \hline
        \textbf{DM} & 1.7423 & 1.6084  & 4.7679  & 3.7639 \\
        \textbf{TPIDM} & 1.7748 & 1.6106  & 4.8928  & 3.7801\\
        \textbf{PIDM} \cite{bastek2024physics} & 1.7309 & 1.5100  & 5.1415  & 4.0328\\
        \textbf{PIDM} \cite{shu2023physics} & 1.8817 & 1.7797 & 5.9067 & 4.0531\\
        \textbf{VAE} & 0.0251 & 0.0188 & 0.0633 & 0.0528\\
        \textbf{AE} & \textbf{0.0228} & \textbf{0.0137} & \textbf{0.0594} & \textbf{0.0464}\\
        \textbf{K-MEANS } & 1.3028 & 1.7321 & 2.6844 & 2.2074\\
        \hline
    \end{tabular}
    \label{table:train-inference-time}
\end{table}
\end{document}